\documentclass[letterpaper, 10 pt, conference]{ieeeconf}  

\IEEEoverridecommandlockouts                              

\usepackage{graphics} 
\usepackage{epsfig} 
\usepackage{mathptmx} 
\usepackage{times} 
\usepackage{amsmath} 
\usepackage{caption}
\usepackage{subfig}
\usepackage{xcolor}
\usepackage{hyperref}
\usepackage{amssymb} 
\usepackage{enumerate}
\usepackage{tikz}

\usepackage[noadjust]{cite}

\usepackage{tikz}

\title{\LARGE \bf
Human-robot collaborative transport personalization via Dynamic Movement Primitives and velocity scaling
}

\author{Paolo Franceschi$^{1,*}$, Andrea Bussolan$^{1,*}$, Vincenzo Pomponi$^{1}$, Oliver Avram$^{1}$, Stefano Baraldo$^{1}$, Anna Valente$^{1}$
\thanks{$^{1}$Department of Innovative Technologies, University of Applied Science and Arts of Southern Switzerland (SUPSI), Lugano, Switzerland
        {\tt\small \{name.surname\}@supsi.ch }}.
\thanks{$^{*}$Paolo Franceschi and Andrea Bussolan equally contributed to the work}
}

\begin{document}

\maketitle
\thispagestyle{empty}
\pagestyle{empty}


\begin{abstract}

Nowadays, industries are showing a growing interest in human-robot collaboration, particularly for shared tasks. 
This requires intelligent strategies to plan a robot's motions, considering both task constraints and human-specific factors such as height and movement preferences. 
This work introduces a novel approach to generate personalized trajectories using Dynamic Movement Primitives (DMPs), enhanced with real-time velocity scaling based on human feedback. 
The method was rigorously tested in industrial-grade experiments, focusing on the collaborative transport of an engine cowl lip section. 
A comparative analysis between DMP-generated trajectories and a standard industrial motion planner (BiTRRT) highlights their adaptability, combined with velocity scaling. 
Subjective user feedback further demonstrates a clear preference for DMP-based interactions. 
Objective evaluations, including physiological measurements from brain and skin activity, reinforce these findings, showcasing the advantages of DMPs in enhancing human-robot interaction and improving user experience.

\end{abstract}




\section{Introduction}

Human-robot interaction (HRI) has emerged as a significant area of research within robotics. 
The ability of robots to interact and collaborate with humans safely, efficiently, and intuitively is crucial for their integration into modern industries \cite{ajoudani2018progress}, opening the era of deliberative robotics \cite{valente2022deliberative}.
A key aspect of HRI is human-aware motion planning, which involves determining the sequence of movements a robot must execute to accomplish a specific task.
Typically, the objective of industrial motion planners is to find a fast and precise motion \cite{7989103}, which requires a trade-off between the two \cite{valente2017smooth}.
Traditional motion planning mainly relies on sample-based methods \cite{doi:10.1177/0278364911406761, 6377468}, which makes the motion of robots sometimes unpredictable, in particular for complex/constrained motions, such as planning in cluttered environments \cite{franceschi2022optimal} or motions in the presence of a human partner \cite{8520620}.
Some recent methods combine online motion re-sampling to consider the human presence \cite{10013661} to modify a pre-planned trajectory according to online situations.
Such methods are effective and efficient in planning collision-free trajectories and are a great tool when the robot and the human do not physically interact.

Specific challenges are faced when dealing with physical Human-Robot Interaction (pHRI) \cite{de2008atlas}, especially in scenarios that require online trajectory modification based on it \cite{losey2017trajectory,10275780,10037763}.
According to human interaction, online modification allows for dynamic obstacle avoidance that might be unknown to the robot \cite{10342014, 10606422} or precise adjustments.
However, they can't guarantee that the modified trajectory is still collision-free with the same certainty offline motion planning guarantees.
Moreover, the trajectories are still not personalized for each specific human user, who may have different requirements due to ergonomy or personal preference.

\begin{figure}[t]
    \centering
    \includegraphics[width=\columnwidth,trim={3cm 5cm 2cm 2cm},clip]{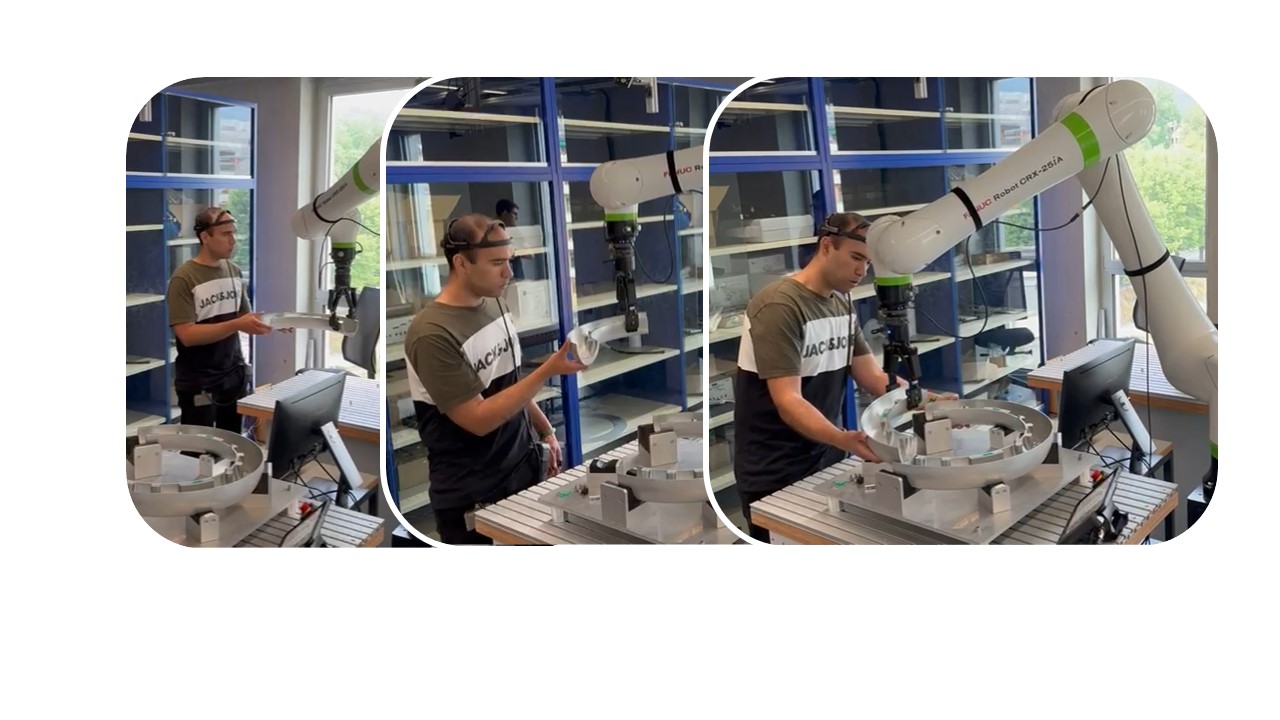}
    \caption{Participant performing the collaborative transport of a section of the engine cowl lip.
    The subject is wearing physiological sensors used to evaluate the stress level during transport. }
    \label{fig: setup}
\end{figure}

For this purpose, Dynamic Movement Primitives (DMPs) \cite{Schaal2006} have been widely used in robotics due to their ability to represent complex motor skills in a simple and generalizable way.
One of the main features of DMPs is that, from a single demonstration, they can generalize to replicate a shape similar to the demonstrated trajectory but with different starting and target positions \cite{ijspeert2013dynamical, 7856889}. 
This particular aspect of DMPs allows trajectory personalization according to each user. At the same time, the trajectory representation by DMPs allows for them to dynamically generate movements that are not bound to the original velocity profile \cite{7030009}.
Even though the method of DMPs was designed to generalize trajectories in free space, they have been applied, over the years, to different tasks involving physical interaction, like obstacle avoidance \cite{zhai2022motion} and iterative hand-guided motion refinement \cite{10136808}.

Given that DMPs provide personalization in the offline motion planning step, we also want to add online modification of the execution speed to make the HRI even more personalized.
Velocity scaling consists of modifying the timing law of a given desired trajectory. 
It is typically used for two primary purposes: (i) to preserve the geometrical path in case the nominal trajectory exceeds the constraints of the system \cite{8477138}, and (ii) for safety purposes in the presence of a human \cite{lippi2018safety,8869047}.
Concerning the second case, the most used approach is the so-called Speed and Separation Monitoring (SSM) \cite{BYNER2019239}, which ensures limited velocity when a human is near the robot.
The first case relies on the joint limits of the robot, while the second scales the velocity according to the relative distance and speed of the human with respect to the robot.
In this work, we propose a velocity scaling law that depends on the physical human-robot interaction measured through the interaction force.

Personalization is a fundamental aspect of HRI in general and pHRI in particular \cite{yang2024}.
It guarantees that the robot behaves in a predictable mode, tailored to each specific user, reducing the risk of high mental and physical loads and injuries.
Therefore, we also want to analyze from a subjective and objective point of view the human responses to the robot's behavior.

In this work, we want to address a co-manipulation task \cite{peternel2018robot}, specifically the collaborative transportation of the section of an inner cowl lip, visible in figure \ref{fig: setup}, by proposing a personalized motion planner based on DMPs.
Moreover, we allow online trajectory velocity scaling according to the physical interaction to make the collaboration even more personalized.
Despite DMPs directly allowing online trajectory modification, we propose to decouple the motion planning and the execution step.
In this way, ensuring that the executed trajectory is collision-free is possible, and the online adaptation is limited to scaling the execution along the predefined path.

Finally, we present a deep evaluation from the human perceptive point of view of the collaborative task. 
We evaluate human preference, both from a physiological and psychological perspective.
We measured Electroencephalography (EEG) and Electrodermal Activity (EDA) and analyzed stress-related metrics relevant to the literature and subjective feelings of fluency during the human-robot interaction using the questionnaire proposed in \cite{hoffman2019}. 

In summary, the main contributions of this work are: 
\begin{enumerate}
  \item personalizing motion planning for a human-robot collaborative task, exploiting DMPs;
  \item a velocity scaling law according to physical Human-Robot Interaction;
  \item a thorough evaluation (objective and subjective) of the pHRI to evaluate human responses.
\end{enumerate}

\section{Method}

This section presents the method adopted in this paper to allow personalized HRI.
First, we recall the fundamental principles of the DMPs.
Then, we present the proposed velocity scaling logic used to speed up and slow down the execution of the trajectory, according to the pHRI.

\subsection{Dynamic Movement Primitives}

In this section, we briefly recall the fundamental principles of the DMPs, as presented in \cite{doi:10.1177/02783649231201196}.
The DMPs used in this work rely on a second-order system described by
\begin{equation}\label{equ:dyn_sys}
    \ddot{y}(t)=\alpha_y(\beta_y(g-y(t))-\dot{y}(t))+f_{DMP}
\end{equation}
where $y(t)$ is the system's state and $\dot{y}(t)$ and $\ddot{y}(t)$ its first and second time derivatives, $\alpha_y$ and $\beta_y$ positive gains, $g$ the goal position, and $f_{DMP}$ is a forcing term.
The forcing term is defined as a linear combination of N nonlinear Radial
Basis Functions (RBFs), computed as 
\begin{equation}\label{equ:f_x}
    f_{DMP}(x) = \frac{\sum_{i=1}^{N} \Psi_i(x) \omega_i }{\sum_{i=1}^{N}  \Psi_i(x)} x(g-y_0),
\end{equation}
where $x$ is the phase variable defined as
\begin{equation}\label{equ:cansys}
    \dot{x} = \alpha_x x,
\end{equation}
with $\alpha_x$ positive gain, $y_0$ is the initial position of the system, $\omega_i$ are weights, $\Psi_i(x)$ are fixed basis functions written as Gaussian functions as follows
\begin{equation}\label{equ:psi}
    \Psi_i(x) = exp(-h_i(x-c_i)^2)
\end{equation}
where $c_i$ are the centers of Gaussian basis functions
distributed along the phase of the movement and $h_i$ their
widths.

The weights $\omega_i$ must be learned, and Locally Weighted Regression (LWR) \cite{atkeson1997locally,schaal1998constructive} is the standard method used for DMPs.
Given a demonstration trajectory $\mathcal{D}=\{y_d(t_0), \;\hdots ,\;y_d(t_f)\}$, \eqref{equ:dyn_sys} is inverted and $f_d$ computed as 
\begin{equation}\label{equ:f_d}
    f_d(t) = \ddot{y_d}(t) - \alpha_y(\beta_y(g-y_d(t))-\dot{y_d}(t)).
\end{equation}
A function approximation problem is formulated to find $\omega_i$ parameters that make $f_d$ as close as possible to $f_{DMP}$.
For each kernel function $\Psi_i$, LWR looks for the corresponding $\omega_i$ that minimizes the locally weighted quadratic error through the following cost function 
\begin{equation}\label{equ:Ji}
    J_i = \sum_{t=1}^P \Psi_i(t)(f_d(t)-\omega_i(x(g-y_0))).
\end{equation}

\subsection{DMP-based trajectory generation}

Consider a robotic manipulator. 
Define with $q=[j_1, \hdots, j_n]^T \in \mathbb{R}^n$ the vector of the robot's joints coordinates, with $j_i$ denoting the $i^{th}$ joint, and $n$ the number of DoFs.
A robot trajectory is defined as a timed sequence of joint configurations as

\begin{equation}\label{equ:traj}
    \tau(t)=[q(t_0)^T,\hdots, q(t_f)^T]^T,
\end{equation}
with $q(t_0)$ and $q(t_f)$ denoting the initial and final configurations, respectively.

In this case, $\tau_d(t)=[q_d(t_0)^T,\hdots, q_d(t_f)^T]^T$ refers to a demonstration trajectory that, in this work, is recorded with a human moving the robot via kinesthetic teaching.
Given $\tau_d(t)$, for each joint $j_i$ is possible to associate a second order system as \eqref{equ:dyn_sys}, and compute \eqref{equ:f_x} to obtain its corresponding DMP.
With the full set of DMPs, generating new trajectories with the same shape of the demonstration $\tau_d(t)$ but different initial and final configurations is possible.

Since the proposed work involves pHRI, which has severe rules to comply with safety standards defined by the normative \cite{ISO10218-1, ISO10218-2, ISOTS15066}, we consider only the path (i.e., the sequence of positions, without velocities and accelerations) and adjust them to comply with speed limitations.
In this way, we have a trajectory with the same shape as the original one computed by the DMPs but with limited velocities that are compliant with safety standards.

Currently, the computed trajectories are not guaranteed to be collision-free, which is out of the scope of this work\footnote{This work focuses on generating and evaluating a personalized trajectory. Also, typical cases of pHRI involve almost-free workspaces where collisions are very unlikely to happen, rather than cluttered environments. To generate a collision-free trajectory, other DMP-based planners exist, such as \cite{zhai2022motion}. }.
Despite this, it is easy to check if there are collisions with known static objects, once the trajectory is computed.
In this work, the trajectory will not be executed if collisions are found.

\subsection{Trajectory parametrization}

Consider now the trajectory to be executed, defined as 
\begin{equation}\label{equ:qd}
    \tau = \tau(s),
\end{equation}
with the variable $s$ determining the path's timing law (i.e., the velocity profile).
Taking its first derivative, 
\begin{equation}\label{equ:dqd}
    \dot{\tau} = \tau^{\prime}(s) \dot{s} = \frac{d \tau(s)}{ds}\frac{ds}{dt}.
\end{equation}
The term $\tau^{\prime}$ is geometrically tangent to curve $\tau$, while the value of the scalar $\dot{s}$ determines the amplitude of vector $\dot{\tau}$. 
Thus, by varying the value of $\dot{s}$, the timing law can be punctually modified while keeping the same geometrical direction.
In discrete time, the update law to select the time parameter $s$ at timestep $k$ is
\begin{equation}\label{equ:scaling}
    s(k+1) = s(k) + \sigma(k) \, \Delta t,
\end{equation}

with $\Delta t$ indicating the time interval between two subsequent computations, and $\sigma$ represents the scaling of the velocity.
When $\sigma=1$, the trajectory is executed at the nominal velocity, as computed by the motion planner, while $\sigma<1$ allows for speed reduction. 
In general, $\sigma>1$ allows the execution of the trajectory to be faster than the nominal one. 
Due to the safety requirements of pHRI, this case is not considered in this work.

\subsection{Velocity scaling}

To make the interaction more comfortable for the human partner, we propose to scale the execution velocity according to human intentions, measured through force interaction.

The velocity vector at the end-effector is given by $v=[v_x,v_y,v_z]^T$, and the force applied by the human is $f=[f_x,f_y,f_z]^T$.
We consider the components of the force along the velocity direction as a measure of the human intention to speed up or slow down the execution of the trajectory.
We define the \textit{agreement parameter} as 
\begin{equation}
    \rho = f \cdot \hat{t},
\end{equation}
where $\hat{t}$ is the versor tangent to the velocity at the current point, computed as $\hat{t}=\frac{v}{||v||}$.

\begin{figure}[h]
    \centering
        \includegraphics[width=0.8\columnwidth,trim={0 0 3cm 0},clip]{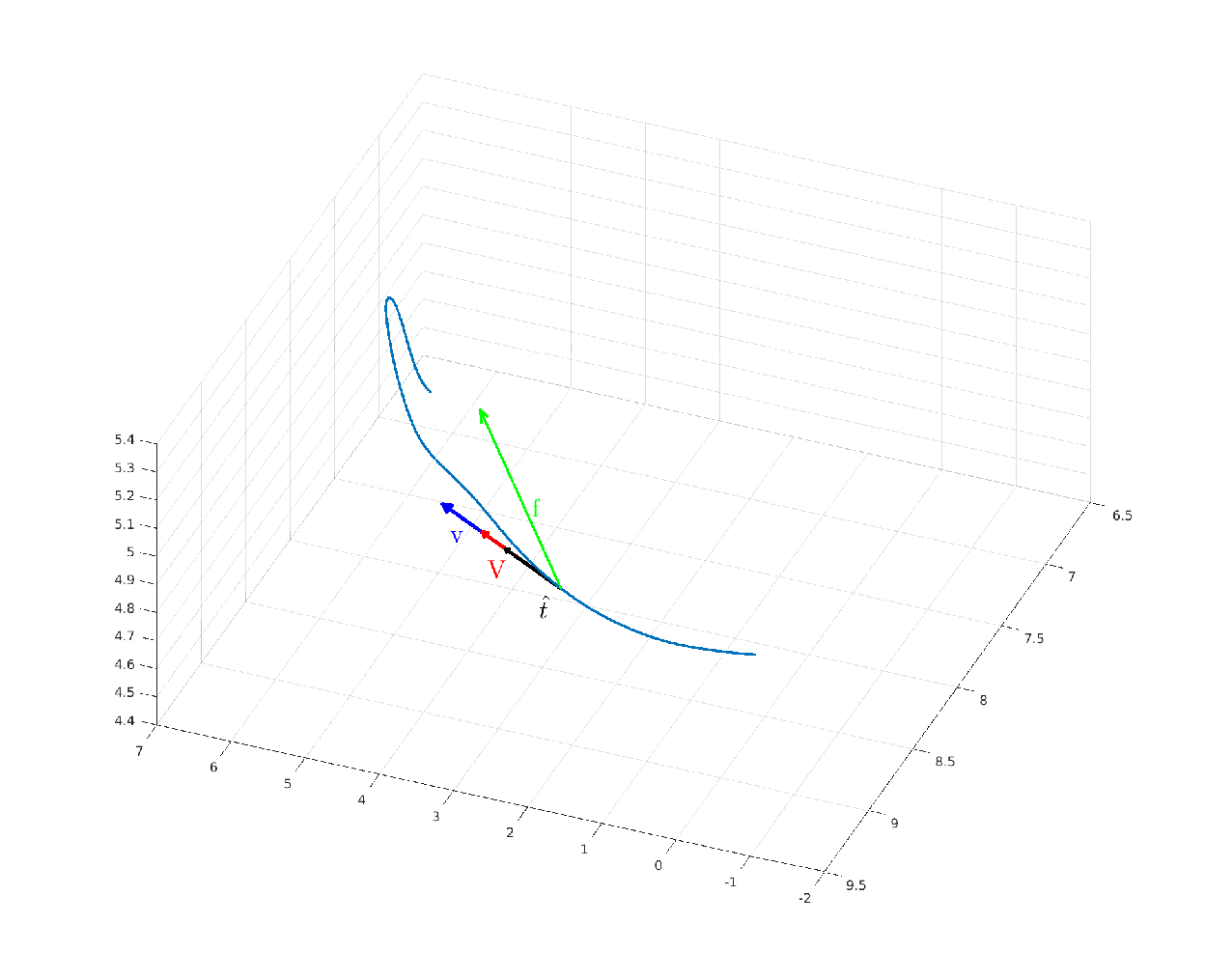}
        \caption{The velocity $v$ and force $f$ vectors, the versor $\hat{t}$, and the vector of force components along the velocity direction.}
  \label{fig:vectors}
\end{figure}

An illustrative example of the components is visible in figure \ref{fig:vectors}.

A sigmoid function is used to convert the agreement parameter $\rho$ into a percentage of velocity scaling for the trajectory as 
\begin{equation}\label{equ:sigmoid}
    \sigma = \frac{1}{1+e^{-m\,\rho}},
\end{equation}
where $m$ is a parameter that defines the slope of the function.
The value of $\sigma$ can then be used in \eqref{equ:scaling} to determine the trajectory execution velocity.

\section{Experiments}

As an experimental scenario to test the proposed method, we selected the collaborative transportation of a long object, namely the section of an engine cowl lip with a length of 0.5 m and a weight of about 6 kg, with the robot grasping it from one side and the human from the other side. The assembly components are based on real aeronautic parts, but have been reproduced with some modifications to avoid IP issues and improve the re-useability in lab experiments.

\subsection{Design of experiments}

The experiment requires the grasping, collaborative transport, and placement of three lip sections.
Before performing the task, the user is asked to teach the robot the transport trajectory using manual guidance.
The trajectory starts above the table where lip sectors are placed and ends above the fixturing, where the three components must be placed. 
The demonstration trajectory does not require precise start and end positions, as the DMPs can receive them at generation time and generalize the demonstrated trajectory accordingly. 
Since the considered type of components to be transported are long and cumbersome (up to 1.5 m, with arc shape), the demonstration trajectory is recorded without the component, and the user grasps the robot directly at its end-effector.

After teaching the trajectory, each user is asked to try it for the transport of one component to check if it is satisfactory or if a new one should be recorded.
When the demonstration trajectory is recorded, each user is asked to perform the task, requiring three collaborative transports (one for each sector of the full lip).

We compare the proposed DMP planning with velocity scaling ($DMP_V$) with the BiTRRT planner \cite{6631158} from the Open Motion Planning Library ($OMPL$) \cite{6377468}.
The experiment is repeated four times for comparison. Two experiments involve DMP planning, and the other two BiTRRT. 
For each planner, in one case, the velocity scaling is active, while in the other, it is not, and the trajectory is executed at the nominal speed. 
Hereafter, we refer to $DMP_V$ and $BiTRRT_V$ for the cases where the velocity scaling is active and to DMP and BiTRRT when the scaling is not active.

A total of 12 users, two female and ten male, aged between 22 and 34 years old, performed the experiments.
The users have different levels of expertise in using robots, ranging from no experience to expert.
To reduce the effect of the learning curve and bias on the evaluation, the order of planning strategy is randomly selected for each user, such that each user performs the task with all the strategies, but in random order.

We want to evaluate subjective preferences for the four strategies in terms of i) geometrical and dynamical properties of the trajectories, ii) subjective feelings, and iii) physiological response.

The experiments are performed on a workcell with a Fanuc CRX-25iAL robot equipped with a Robotiq FT300 force/torque sensor mounted on the wrist, which is used to read the force for velocity scaling, and a Robotiq 2F-140 gripper for grasping the piece. The experimental setup with a user performing the task is visible in figure \ref{fig: setup}.

\subsection{Performance indexes}

To evaluate the personal responses of the users, we propose different evaluation metrics.
First, we want to verify if the executed trajectories present features that highlight whether the velocity scaling and path personalization are actually useful.
Second, we want to evaluate subjective user responses with a standard questionnaire.
Finally, we use various physiological measurements and extract performance indexes related to unconscious subjective feelings during the transport.

For each of the proposed evaluations, we are specifically interested in the collaborative transportation phase. Therefore, all the results are relative to the transport only.

\subsubsection{Trajectory indexes}

We measure trajectory features to understand the differences between the BiTRRT and DMP trajectories. 
First of all, we measure the execution time, to understand if velocity scaling is useful to make the subjects' perception of the collaborative operation more pleasant.
In particular, we evaluate, for each transport modality, the execution time averaged on subjects $i=1,...,n$, defined as 

\begin{equation}\label{equ:time_avg}
    T_{avg,mode}=\frac{1}{n}\sum_{i=1}^n{T_{i,mode}},
\end{equation}
where $mode=\{DMP,BiTRRT,DMP_V,BiTRRT_V\}$ denotes the modality proposed for the transport.

We also want to evaluate how different subjects teach the trajectories in relation to their height, as an indicator of the improvment in ergonomy provided by the proposed approach. To this aim, we compute for each subject the ratio between the average height of the executed trajectories and the user's own height, i.e., 
\begin{equation}\label{equ:height_avg}
    R_{height,mode}=\frac{ Z_{avg,mode} }{U_{height}},
\end{equation}
where $U_{height}$ denotes the specific user height, and $Z_{avg,mode}$ denotes the average height of execution for each user trajectory.
In this case, the result is independent of velocity scaling, and all the DMPs and BiTRRT are used to compute the respective averages. Therefore, $mode=\{DMP, BiTRRT\}$ and $DMP$ and $BiTRRT$ include both versions of the transport, with and without velocity scaling.

\subsubsection{Questionnaire}

After each experiment involving the transport of all three components, users are asked to answer a questionnaire.
The questionnaire is based on \cite{hoffman2019}, with additional questions specific to the proposed use case.
In particular, in \cite{hoffman2019}, the author proposes different scales under which different questions are aggregated. 
We choose only the scales that are relevant to our use case, focusing on the effect that personalized trajectories have on Human-Robot Fluency, Robot Relative Contribution, Trust in the Robot, and Positive Teammate Traits. 
We also singled out the question \textit{The robot was cooperative}, even though it is not included in the aforementioned aggregated metrics, because of its relevance to our scenario.
To directly investigate the participants' preferences, they were asked to express their preferences regarding the robot's velocity and predictability with the additional questions: \textit{The robot trajectory was predictable} and \textit{The robot velocity was adequate}. Moreover, at the end of the experiments, the users are asked to compile their personal ranking of the four modalities proposed for transport. 
We then scored the answers from 0 (less preferred) to 3 (most preferred) and computed an aggregate ranking.

\subsubsection{Physiological indexes}
To assess the cognitive load and stress involved in the tested transport modes, given the brevity of the task in consideration, we opted for the use of EEG and EDA signals, which rapidly display the physiological changes over external stimuli, compared to other signals like ECG. In particular, we focus on the mean value of the Skin Conductance Level ($SCL$) and on the theta-to-alpha power ratio of the EEG, which are shown to be useful to detect cognitive load and stress.

In \cite{posada2018, setz2010} the value of the $SCL$ is used to discriminate between rest conditions and psychological stress conditions, where in the latter the $SCL$ significantly increases. 
The $SCL$ index is computed as the mean value of the signal during the collaborative transport phases of the task. 
The EEG's theta-to-alpha power ratio is computed as the average between the ratios over the right and left hemispheres ($\theta_{F3}$ / $\alpha_{P3}$, and $\theta_{F4}$ / $\alpha_{P4}$). 
In \cite{shimomura2008, raufi2022}, a higher level of this ratio is found to be significant to discriminate between levels of mental workload.

To measure such signals, the participants are equipped with an EEG headset\footnote{https://www.bitbrain.com/neurotechnology-products/dry-eeg/diadem} and EDA sensor\footnote{https://www.bitbrain.com/neurotechnology-products/biosignals/versatile-bio} on the index and middle fingers of the non-dominant hand.
A user wearing the sensors is shown in Fig. \ref{fig: setup}.
Before the experiment, users are first asked to remain seated for 2 minutes, recording a rest baseline. 
Then, participants are asked to transport and assemble the three pieces cooperatively.

The EDA signal was filtered using a $4^{th}$ order low-pass filter coupled with a convolutional signal smoothing. 
The EEG signal was filtered using a $2^{nd}$ order band-pass filter with a frequency range from $0.5$ to $40$ Hz, coupled with a notch filter with a frequency range from $49$ to $51$ Hz, to remove the power-line background noise. We compute $\alpha$ waves ([8, 13] Hz) and $\theta$ waves ([4, 8] Hz) power using Welch's Power Spectral Density (PSD) \cite{welch1967}.
Welch’s method estimates the power spectrum of a signal by segmenting it into overlapping windows, computing the Discrete Fourier Transform (DFT) for each window, and then averaging the squared magnitudes. The PSD is computed as follows:

\begin{equation}
    P(\omega) = \frac{1}{K} \sum_{k=1}^{K} \frac{|X_k(\omega)|^2}{M},
\end{equation}

where $X_k(\omega)$ is the DFT of the k-th windowed segment, and M is the number of points in each segment.
Given the high inter-subject variability, we normalized the evaluated metrics using a subject-specific min-max normalization. 

\section{Results}

This section analyzes the results of the presented indexes.

\subsection{Trajectory indexes}

\subsubsection{Time}

Table \ref{tab:time_comparison} 
shows the average and standard deviations of the execution time of the four proposed modalities.
If velocity scaling is enabled, the average time rises by about $40\%$ and $35\%$ compared to the nominal execution time for the BiTRRT and DMP, respectively.
This indicates that, in general, users might need to slow down the execution in some particular circumstances.

\begin{table}[h]
    \centering
    \begin{tabular}{|c|c|c|c|c|}
    \hline
            & $DMP$ & $BiTRRT$ & $DMP_V$ & $BiTRRT_V$\\
    \hline
      times [s] & 9.0 $\pm$ 1.3 & 9.7 $\pm$ 2.3 & 12.1 $\pm$ 2.7 & 13.7 $\pm$ 2.2 \\
    \hline
    \end{tabular}
    \caption{Mean and standard deviation of the transport duration for the different cases, as from \eqref{equ:time_avg}.}
    \label{tab:time_comparison}
\end{table}

\begin{figure}[h]
    \centering
        \includegraphics[width=\columnwidth]{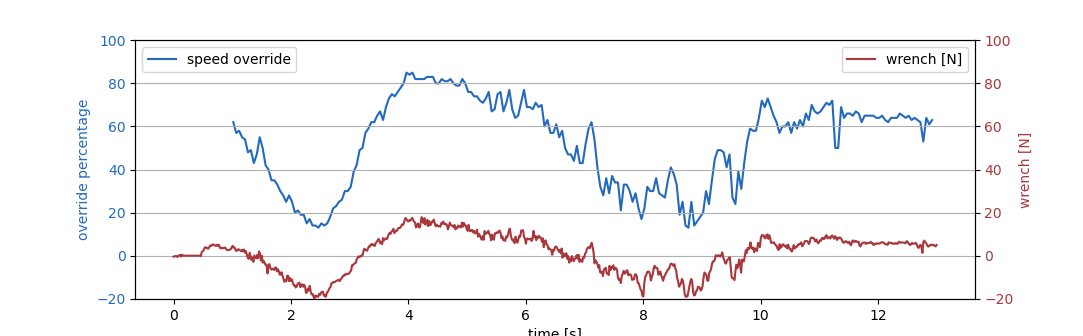}
        \caption{Comparison between aligned wrench and speed override.}
  \label{fig:speed_wrench}
\end{figure}

As an example, Fig. \ref{fig:speed_wrench} shows the speed override percentage compared to the wrench components applied along the velocity direction.
It can be noticed that two times, in the intervals 1-3 and 7-10 seconds, the user decided to slow the execution down.
These two intervals correspond to two specific phases of the execution.
The first happens at the beginning of the motion, when the user gets ready for the transport. 
The second corresponds to a delicate phase of the transport, when the lip approaches the fixturing for its final placement.
In this phase, slowing down the robot allows adjusting the user's position around the table and the part's position over the fixturing.
The fact that the velocity scaling method allows more pleasant execution is also confirmed by an explicit question of the questionnaire, as shown in Figure \ref{fig: histogram}.


\subsubsection{Height}

\begin{figure}[h]
    \centering
        \includegraphics[width=\columnwidth]{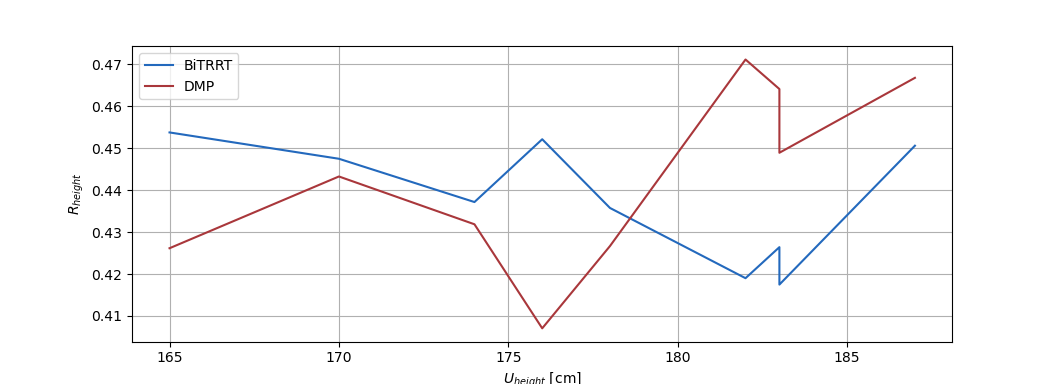}
        \caption{Ratio between average trajectory height and users' heights, for each user in the study.}
  \label{fig:avg_z}
\end{figure}

Figure \ref{fig:avg_z} shows the variation of the ratio between average trajectory height and users' heights.
It can be noticed that higher average trajectories correspond to taller users, showing that users' height plays a role in the personalization of the interaction.
On the contrary, BiTRRT, by proposing sample-based trajectories, cannot take users' height into account, and the ratio does not describe adaptation and personalization.
Moreover, one user preferred a low trajectory even if it was taller than other users (176 cm).
This indicates that user height, in general, relates directly to higher trajectories, but this rule does not apply to everyone.
In this sense, it cannot be taken as a general constraint for trajectory generation by other methods, making the use of the proposed DMP a valuable solution in terms of personalized path generation.

\begin{figure}[h]
\centering

\begin{minipage}[b]{0.48\columnwidth}
  \centering
  \begin{tikzpicture}
    \node[anchor=south west, inner sep=0] (img1) at (0,0)
      {\includegraphics[width=1.1\linewidth,trim={7cm 3cm 5cm 2cm},clip]{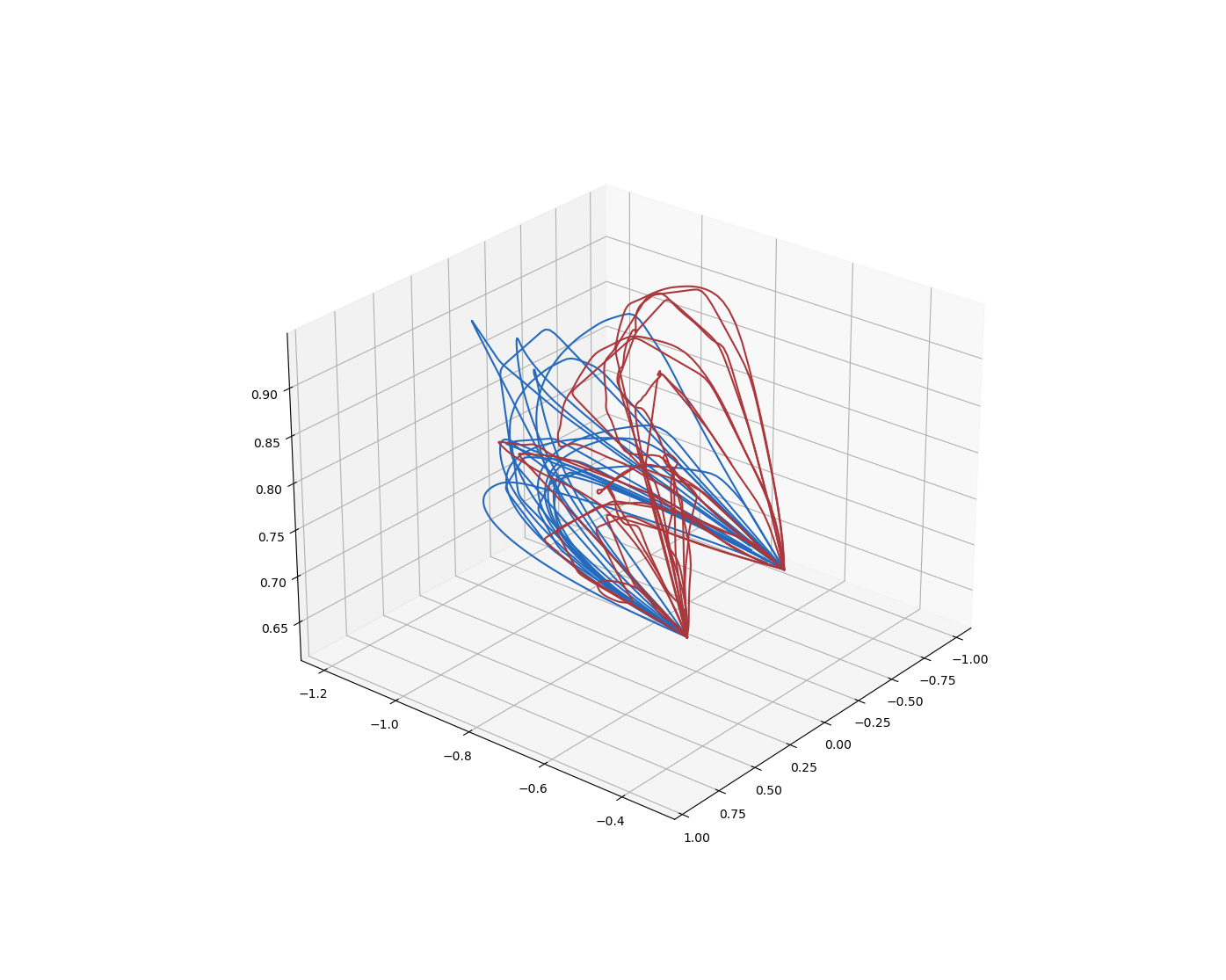}};
    \begin{scope}[x={(img1.south east)}, y={(img1.north west)}]
\draw[->, red, thick]   (0.2,0.35) -- ++({0.08*cos(-135},{0.08*sin(-135)}) node[above left] {$x$};
\draw[->, green, thick] (0.2,0.35) -- ++({-0.08*sin(-135)},{0.08*cos(-135)}) node[above right] {$y$};
\draw[->, blue, thick]  (0.2,0.35) -- ++(0,0.08) node[above] {$z$};
    \end{scope}
  \end{tikzpicture}
  \caption*{(a) All trajectories}
\end{minipage}
\hfill
\begin{minipage}[b]{0.48\columnwidth}
  \centering
  \begin{tikzpicture}
    \node[anchor=south west, inner sep=0] (img2) at (0,0)
      {\includegraphics[width=\linewidth,trim={7cm 3cm 5cm 2cm},clip]{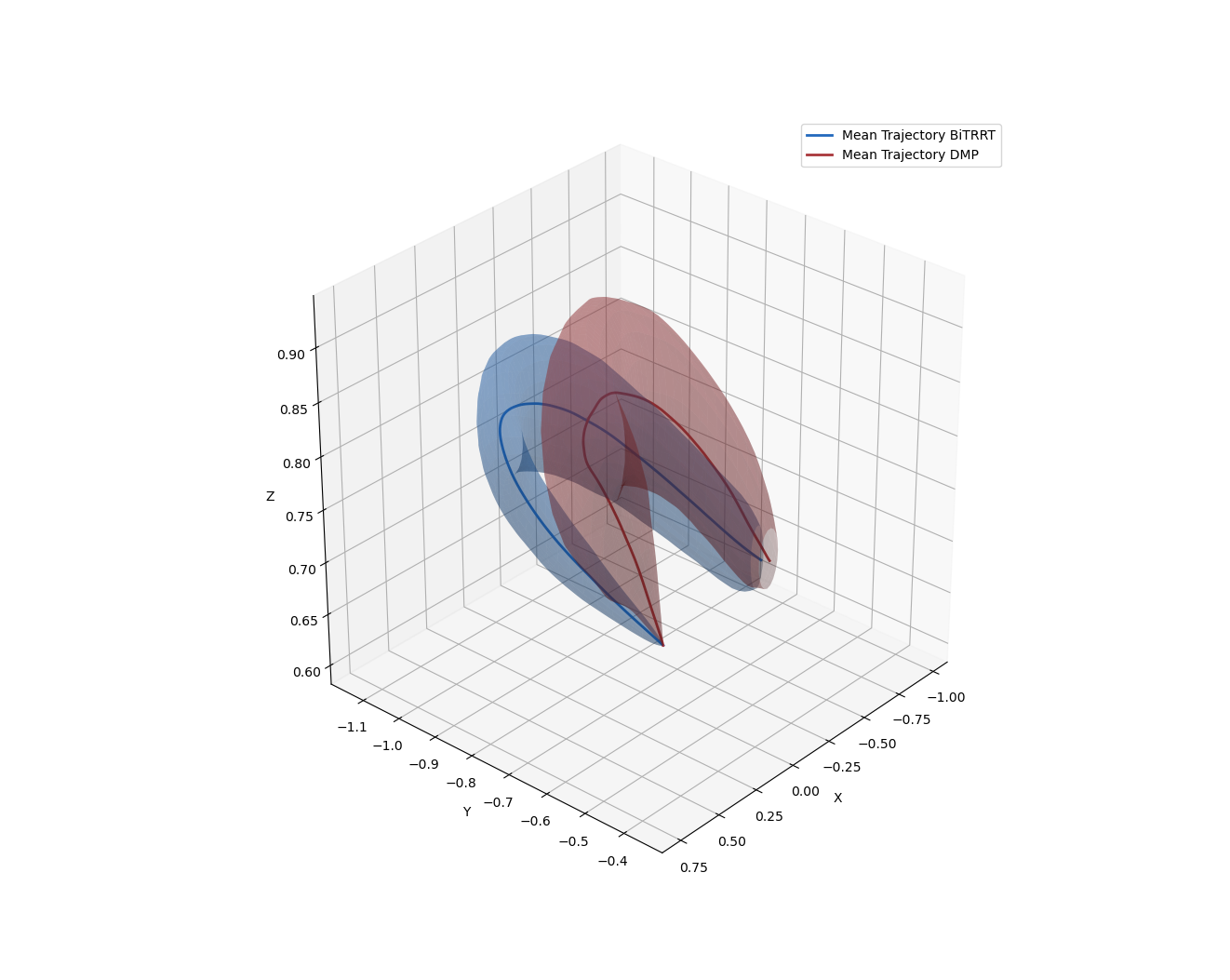}};
    \begin{scope}[x={(img2.south east)}, y={(img2.north west)}]
\draw[->, red, thick]   (0.2,0.35) -- ++({0.08*cos(-135},{0.08*sin(-135)}) node[above left] {$x$};
\draw[->, green, thick] (0.2,0.35) -- ++({-0.08*sin(-135)},{0.08*cos(-135)}) node[above right] {$y$};
\draw[->, blue, thick]  (0.2,0.35) -- ++(0,0.08) node[above] {$z$};
    \end{scope}
  \end{tikzpicture}
  \caption*{(b) Average trajectories}
\end{minipage}

\vspace{1ex}
\caption{Average (right) and all trajectories (left) planned by the BiTRRT (in blue) and DMP (in red) for the assembly of the first component. X-Y-Z units are in meters.}
\label{fig:trajectory_comparison}
\end{figure}

Figures \ref{fig:trajectory_comparison}-a and \ref{fig:trajectory_comparison}-b show the trajectories and their averages and standard deviations, respectively, computed by BiTRRT and DMP for the collaborative transport of one lip sector.
The trajectories computed by DMP based on the demonstration are higher and closer to the robot's base, showing that users, in general, prefer to move straight to the goal at a higher level.

\subsection{Questionnaire}

\begin{figure}[bt]
    \centering
    \includegraphics[width=0.95\columnwidth, keepaspectratio]{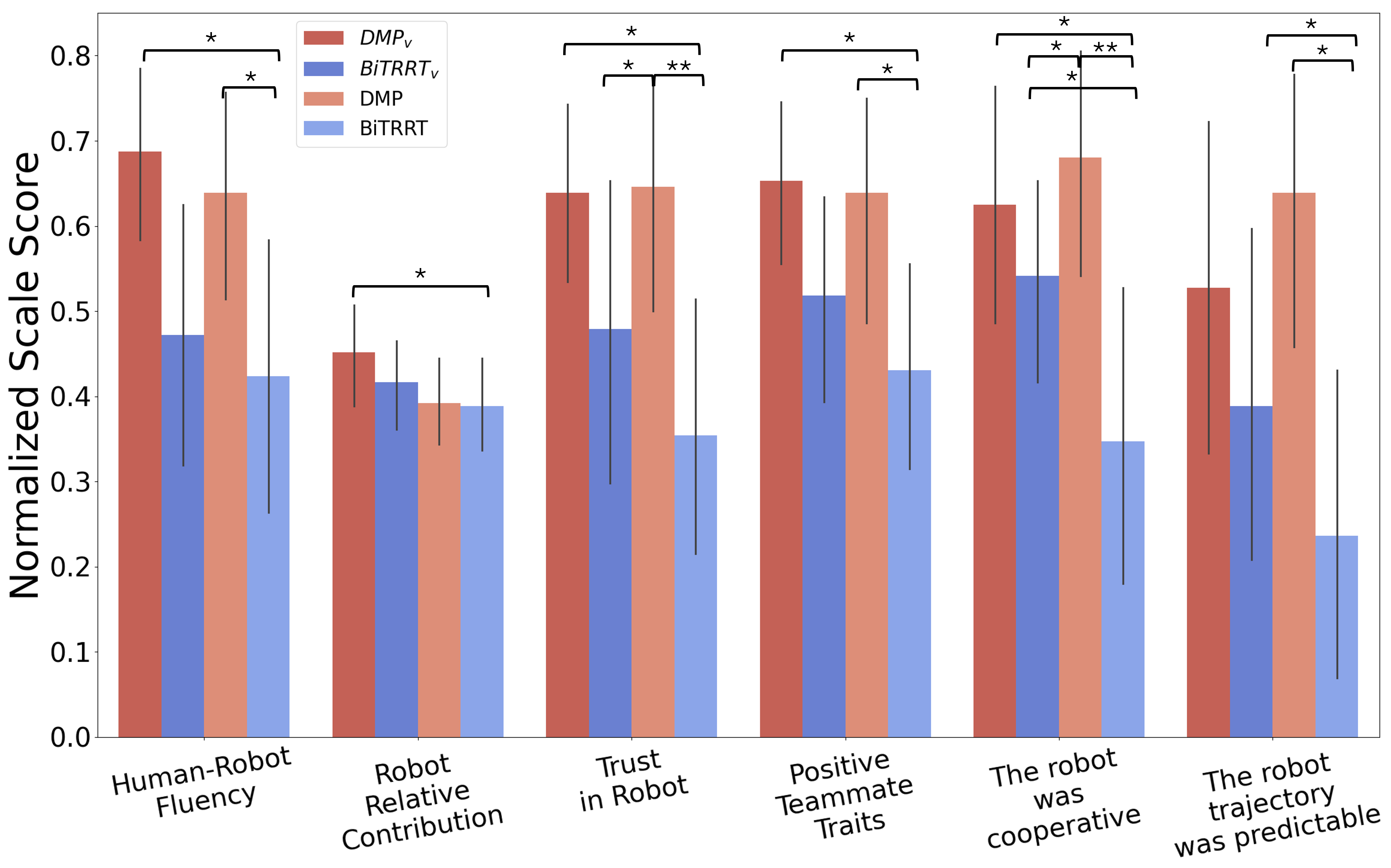}
    \caption{Normalized scale scores for different trajectory generation methods across various human-robot interaction metrics.}
    \label{fig: barplots}
\end{figure}

Figure \ref{fig: barplots} represents a comparative analysis of subjective ratings on the selected HRI scales and questions. The represented scores are normalized with respect to the maximum achievable scores. Significant differences between conditions are highlighted with p-value annotations ($p < 0.05: *$, $p < 0.01: **$).

The $DMP$-based conditions consistently achieve higher ratings in \textit{Human-Robot Fluency} and  \textit{Trust in Robot} compared to $BiTRRT$ conditions. This suggests that $DMP$ allow for smoother and more predictable robot motion, which enhances the user's perception of fluency and trust. The increased trust could be attributed to the deterministic nature of $DMP$-generated trajectories, as opposed to $BiTRRT$, which may introduce variability due to its sampling-based nature, which is reflected on the lower ratings for \textit{The Robot Trajectory was Predictable} in $BiTRRT$ conditions. 
Moreover, improved \textit{Positive Teammate Traits} ratings indicate that participants viewed the robot as more competent, reliable, and cooperative when using $DMP$. It can be noticed that participants perceived the robot as more cooperative if they could modify the robot's speed, suggesting that incorporating velocity adaptation further enhances the interaction experience by allowing the robot to adjust its motion in response to human actions, leading to improved synchrony and cooperation.

\begin{figure}[bt]
    \centering
    \includegraphics[width=0.95\columnwidth, keepaspectratio]{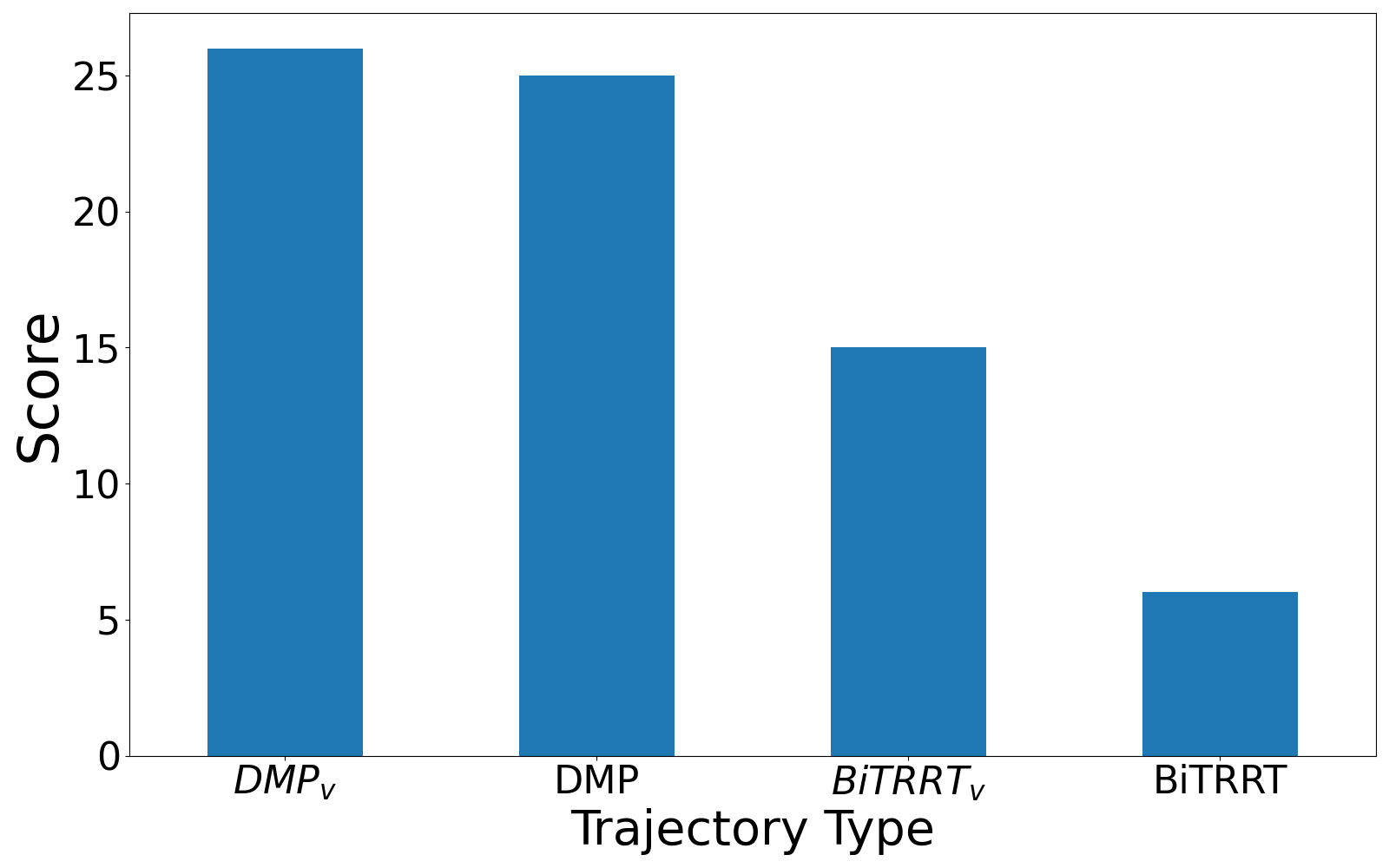}
    \caption{Overall ranking scores for different trajectory generation methods, illustrating their comparative performance.}
    \label{fig: ranking}
\end{figure}

Figure \ref{fig: ranking} represents the overall ranking for the different transport types. These scores indicate an overall preference for $DMP$-based methods over $BiTRRT$-based ones.

\begin{figure}[bt]
    \centering
    \includegraphics[width=0.95\columnwidth, keepaspectratio]{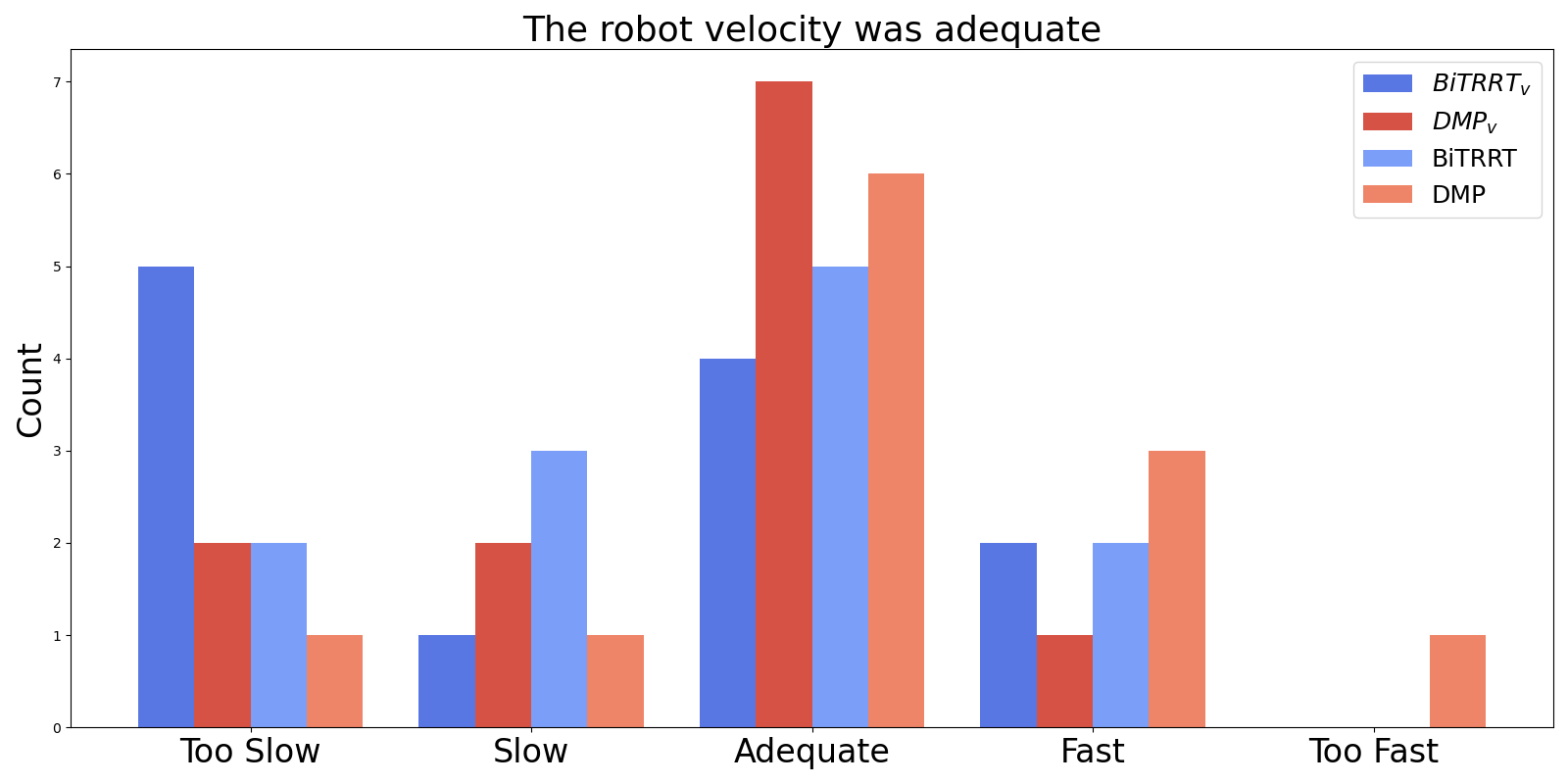}
    \caption{Bar plot of user preferences regarding robot velocity across different trajectory generation methods.}
    \label{fig: histogram}
\end{figure}

Additionally, Figure \ref{fig: histogram} shows the bar plots of the answer to \textit{The robot velocity was adequate}. The highest \textit{Adequate} ratings are given to the $DMP_{v}$ method, suggesting that this approach leads to an optimal user experience. The higher \textit{Too Slow} and  \textit{Slow} ratings in $BiRTTR$ suggest that default path planning velocities may not align well with user expectations in this type of collaborative task.

\subsection{Physiological indexes}

\begin{figure}[h]
\centering
\subfloat[][EDA signals of one participant during the different transports.\label{fig: eda_ts}]{
        \includegraphics[width=0.48\columnwidth, keepaspectratio]{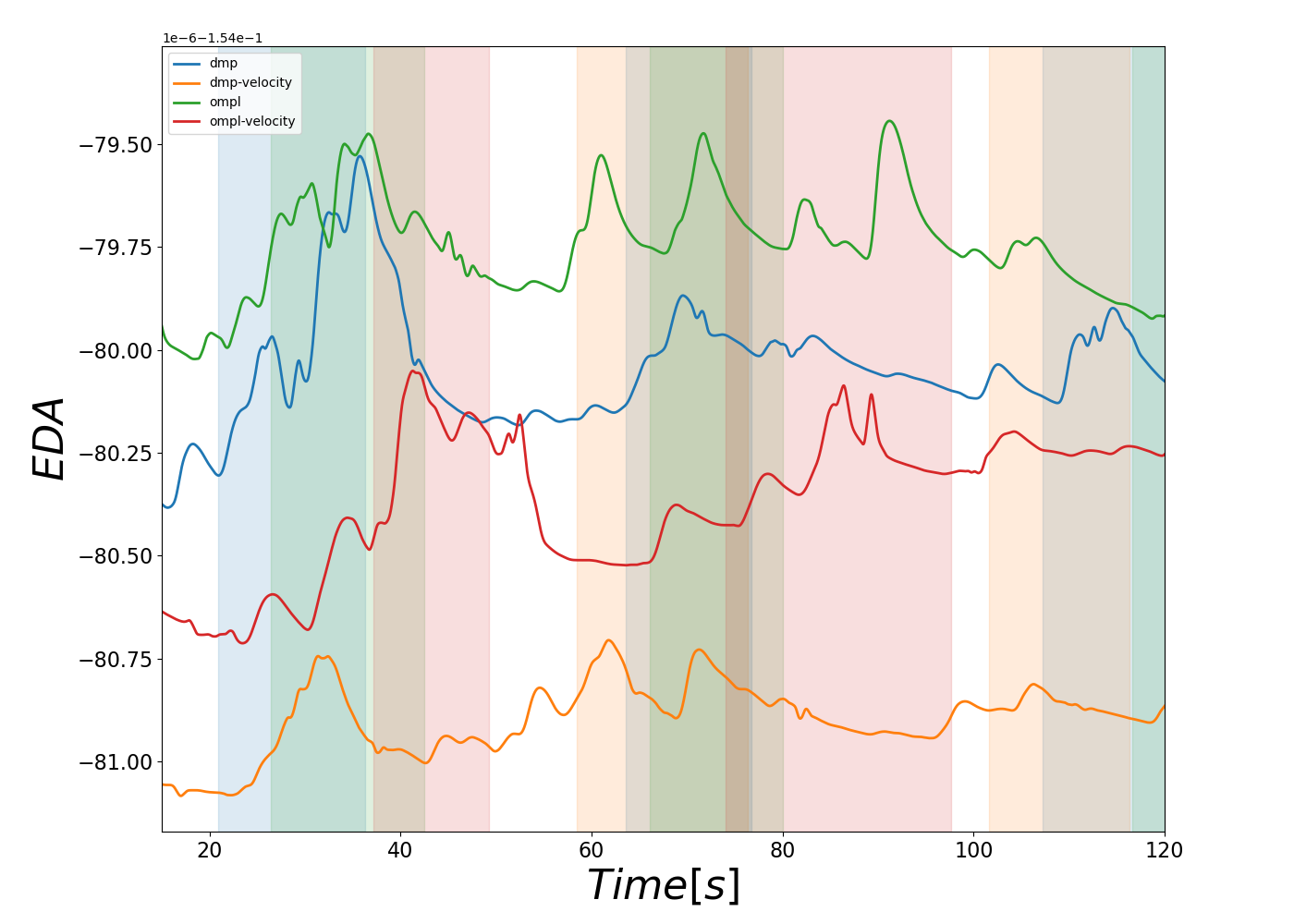}}
     \hfill
\subfloat[][Boxplot of the mean skin conductance level.\label{fig: eda_mean}]{
        \includegraphics[width=0.48\columnwidth, keepaspectratio]{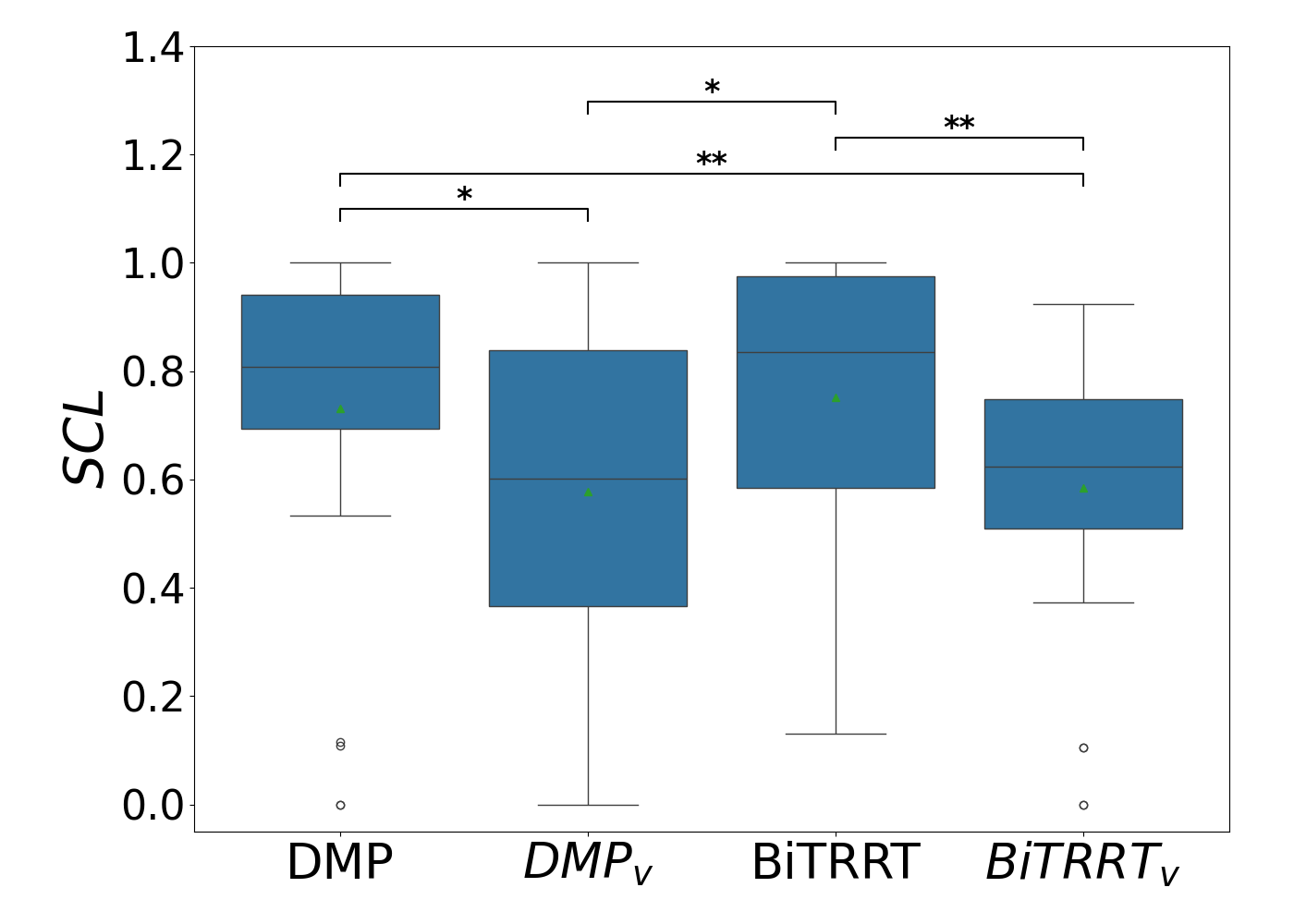}}
     \hfill
        \caption{Results obtained from the Electrodermal Activity.}\label{fig: eda}
\end{figure}

\begin{figure}[t]
    \includegraphics[width=\columnwidth]{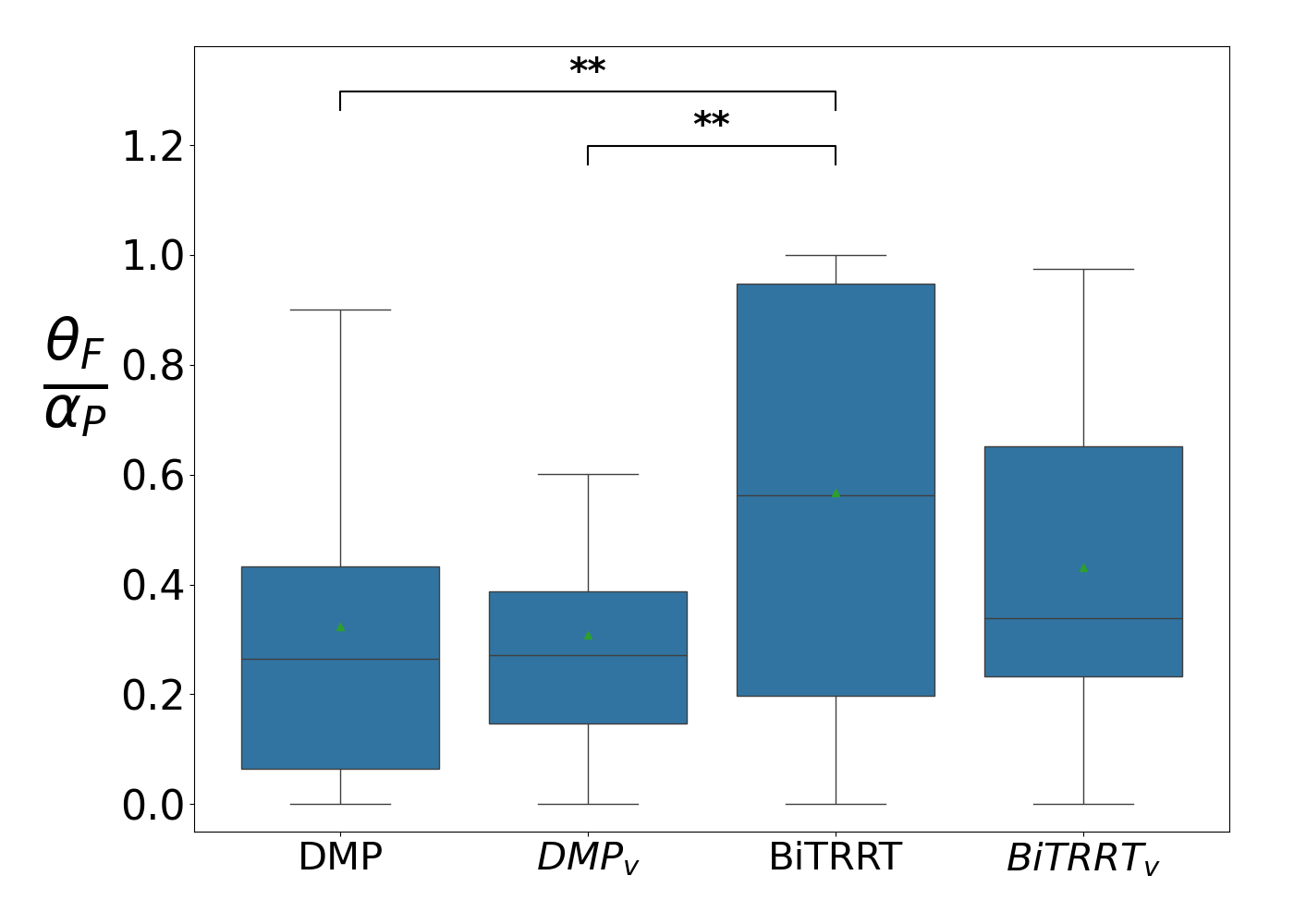}
    \caption{Boxplot of the ratio between frontal $\theta$ power and parietal $\alpha$ power. Outliers are not shown for clarity.}
    \label{fig: eeg_ratios}
\end{figure}

\begin{figure}[t]
    \centering
    \begin{minipage}{0.99\columnwidth}
        \centering
        \subfloat[][$DMP_{v}$ transport\label{fig:eeg1}]{
            \includegraphics[width=0.48\textwidth]{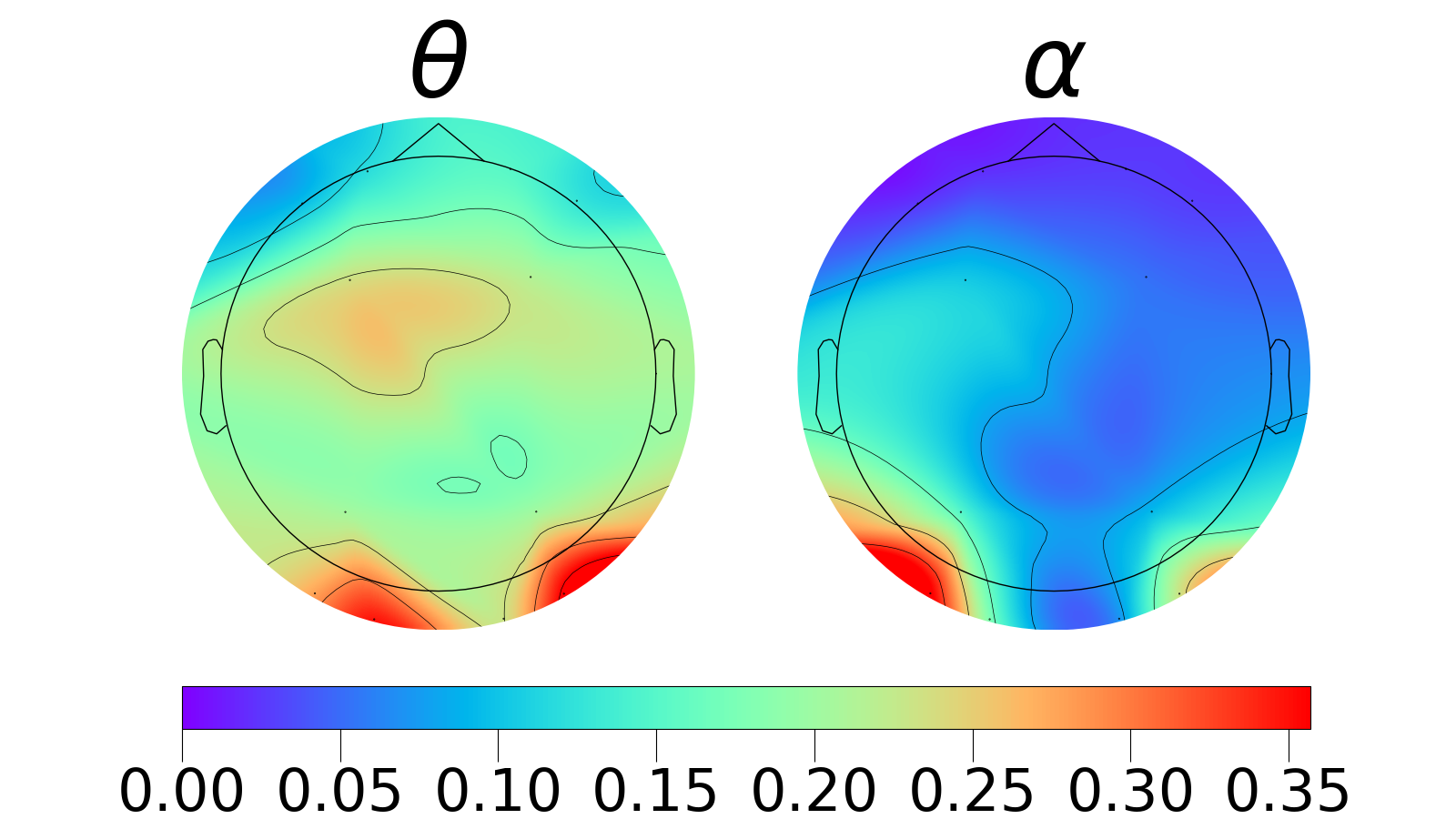}
        }
        \hfill
        \subfloat[][BiTRRT transport\label{fig:eeg2}]{
            \includegraphics[width=0.48\textwidth]{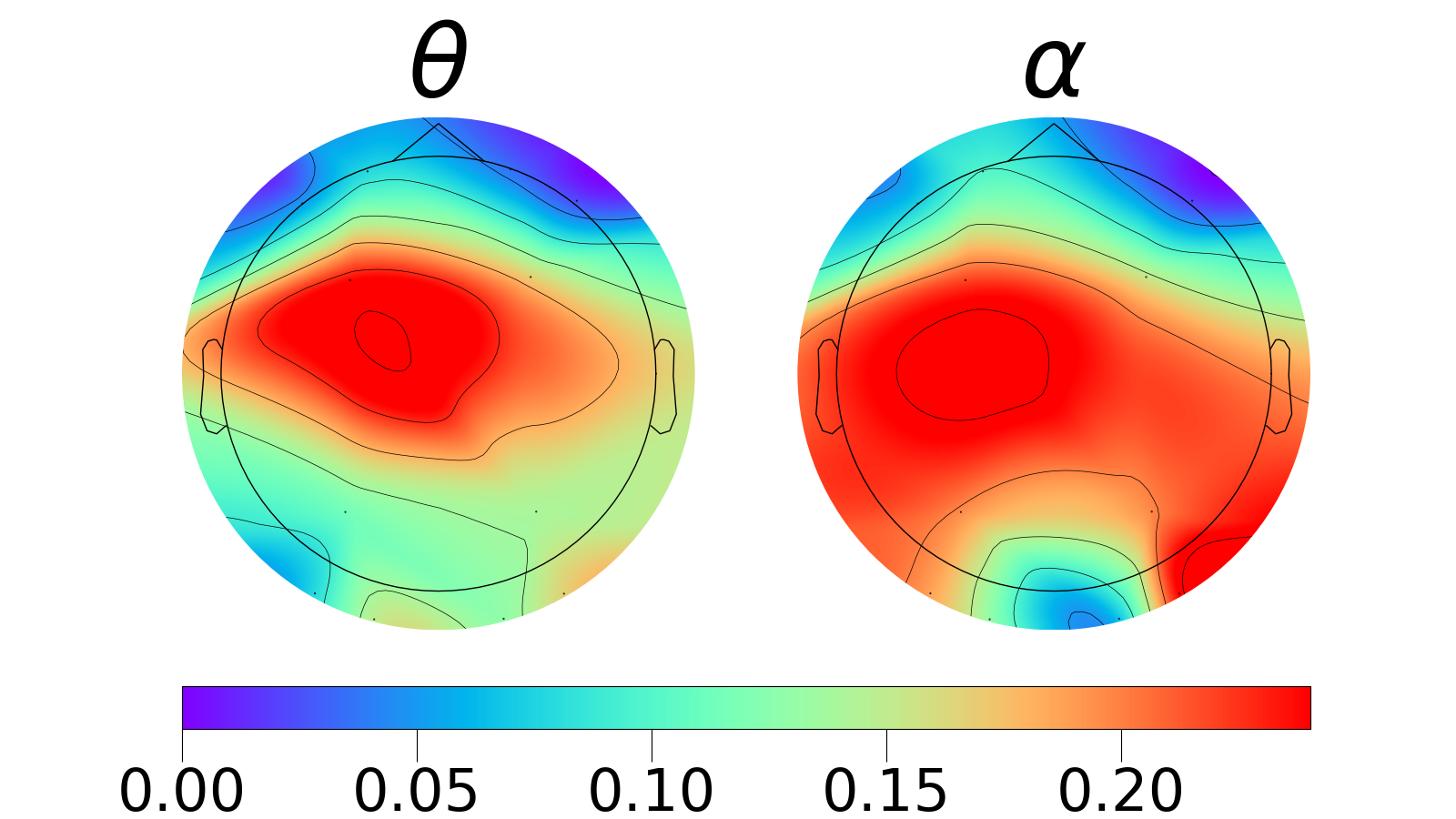}
        }
        \end{minipage}
    \caption{$\alpha$ and $\theta$ powers distribution during two different transports. The colormap indicates the integrated spectral power for the specific frequency bands. Unit of spectral power is $\mu V^{2}$.}
    \label{fig: head}
\end{figure}



Considering EDA metrics, we tested within-group normality using the Shapiro-Wilk test, which assesses whether the data follow a normal distribution. 
Since normality was not satisfied, we tested the difference for each pair of transport types using the Mann-Whitney U rank test, a non-parametric alternative to the t-test that compares two independent groups. The results highlight a significant increase in $SCL$ during the fixed-velocity transports, indicating a higher activity of the sympathetic nervous system, which correlates with higher stress and/or cognitive load. The boxplot in Fig. \ref{fig: eda_mean} represents the distribution of the average EDA values across transport types, while Fig. \ref{fig: eda_ts} shows the time series of one participant's signal, where the higher values of the EDA mean during fixed velocity transports are evident.

Regarding the EEG metrics, after confirming the normality was not satisfied with the Shapiro-Wilk test, we performed paired Mann-Whitney U rank tests. 
The results, shown in Fig. \ref{fig: eeg_ratios}, indicate a significant increment in the mean of the theta-to-alpha ratio, which is related to an increase in the mental workload during the BiTRRT-based transports. 
An example is shown in Fig. \ref{fig: head}, where the increase of $\theta$ power in the F3 and F4 electrodes, and the simultaneous decrease in the $\alpha$ waves power in the P3 and P4 electrodes, are clear.

Therefore, qualitative and quantitative metrics suggest that users experienced a more fluent collaboration using personalized trajectories, which is also reflected by an improvement over physiological metrics, which are considered indicators of mental load.

\section{Conclusions}

This work proposes using DMPs for the offline planning of a personalized trajectory for human-robot collaborative transport and online velocity scaling according to the human-robot interaction force, to allow safe and smooth interaction.
We propose to decouple the planning and the scaling tasks, ensuring that the personalized trajectory can be computed/checked to be collision-free. 
Still, the user can personalize its dynamics during online transport, avoiding deviations from the collision-free path.
We tested the proposed approach with experiments on the collaborative transportation of three sections of an engine cowl lip to be assembled with several participants.
We compared the proposed method with a state-of-the-art motion planner, the BiTRRT.
Results show that with the proposed method, the user's preferences regarding geometrical path and speed execution are respected, compared to standard sample-based motion planning algorithms.
Moreover, we evaluated the users' preferences in terms of subjective feelings, perceived interaction, and objective physiological signals.
The questionnaire shows that, in general, the interaction is perceived as more fluent and pleasant when using the proposed approach, and the robot appears more trustworthy.
The analysis of the physiological signals confirmed that, even from an objective evaluation, the proposed collaborative transport method is less stressful than alternatives based on standard motion planning.

Future works will develop a more general motion planner capable of adapting to the user's preferences and different tasks.
Moreover, the physiological signals will be used to modify the trajectory generation and execution according to the detected stress level.

The \textit{ros2\_fanuc\_interface} is available here 
\url{https://github.com/paolofrance/ros2_fanuc_interface}, and the trajectory scaling controller here \url{https://github.com/paolofrance/scaled_fjt_controller}.

\section*{ACKNOWLEDGMENT}
The research has been funded by the EU project FLUENTLY. Grant agreement no 101058680.



\end{document}